\documentclass[conference]{IEEEtran}
\IEEEoverridecommandlockouts
\usepackage{cite}
\usepackage{amsmath,amssymb,amsfonts}
\usepackage{algorithmic}
\usepackage{subfigure} 
\usepackage{graphicx}
\usepackage{tabularx}
\usepackage{textcomp}
\usepackage{multirow} 
\usepackage{xcolor} 
\usepackage{xcolor}
\usepackage{todonotes} 

\def\BibTeX{{\rm B\kern-.05em{\sc i\kern-.025em b}\kern-.08em
    T\kern-.1667em\lower.7ex\hbox{E}\kern-.125emX}}
\begin{document}

\title{Investigating Traffic Accident Detection Using Multimodal Large Language Models%
\thanks{\textcolor{red}{This work has been accepted for presentation at the 2025 IEEE International Automated Vehicle Validation Conference (IAVVC 2025). 
The final version will be available at IEEE Xplore.}}}

\author{\IEEEauthorblockN{1\textsuperscript{st} Ilhan Skender}
\IEEEauthorblockA{\textit{Embedded Systems Group (Dept.-E)} \\
\textit{Virtual Vehicle Research GmbH}\\
Graz, 8010 Austria \\
ilhan.skender@v2c2.at}
\and
\IEEEauthorblockN{2\textsuperscript{nd} Kailin Tong}
\IEEEauthorblockA{\textit{Control Systems Group (Dept.-E)} \\
\textit{Virtual Vehicle Research GmbH}\\
Graz, 8010 Austria \\
0000-0002-7040-9237}
\and
\IEEEauthorblockN{3\textsuperscript{rd} Selim Solmaz}
\IEEEauthorblockA{\textit{Control Systems Group (Dept.-E)} \\
\textit{Virtual Vehicle Research GmbH}\\
Graz, 8010 Austria \\
selim.solmaz@v2c2.at}
\and
\IEEEauthorblockN{4\textsuperscript{th} Daniel Watzenig}
\IEEEauthorblockA{\textit{Institute of Visual Computing} \\
\textit{Graz University of Technology and}\\
\textit{Virtual Vehicle Research GmbH}\\
Graz, 8010 Austria \\
daniel.watzenig@tugraz.at}
}

\maketitle

\begin{abstract}

Traffic safety remains a critical global concern, with timely and accurate accident detection essential for hazard reduction and rapid emergency response. Infrastructure-based vision sensors offer scalable and efficient solutions for continuous real-time monitoring, facilitating automated detection of accidents directly from captured images. This research investigates the zero-shot capabilities of multimodal large language models (MLLMs) for detecting and describing traffic accidents using images from infrastructure cameras, thus minimizing reliance on extensive labeled datasets.
Main contributions include: (1) Evaluation of MLLMs using the simulated DeepAccident dataset from CARLA, explicitly addressing the scarcity of diverse, realistic, infrastructure-based accident data through controlled simulations; (2) Comparative performance analysis between Gemini 1.5 and 2.0, Gemma 3 and Pixtral models in accident identification and descriptive capabilities without prior fine-tuning; and (3) Integration of advanced visual analytics, specifically YOLO for object detection, Deep SORT for multi-object tracking, and Segment Anything (SAM) for instance segmentation, into enhanced prompts to improve model accuracy and explainability.
Key numerical results show Pixtral as the top performer with an F1-score of 71\% and 83\% recall, while Gemini models gained precision with enhanced prompts (e.g., Gemini 1.5 rose to 90\%) but suffered notable F1 and recall losses. Gemma 3 offered the most balanced performance with minimal metric fluctuation. These findings demonstrate the substantial potential of integrating MLLMs with advanced visual analytics techniques, enhancing their applicability in real-world automated traffic monitoring systems.
\end{abstract}


\section{Introduction}


Recent advancements in artificial intelligence (AI), particularly the rise of multimodal large language models (MLLMs), hold significant promise for enhancing the safety of Intelligent Transportation Systems (ITS). Infrastructure-based vision sensors provide a scalable and efficient solution for real-time monitoring, enabling automated incident detection and description from visual data. Leveraging the zero-shot capabilities of MLLMs, these systems can accurately identify incidents without requiring extensive prior training.
Meanwhile, Connected and Automated Vehicles (CAVs) further contribute to road safety by integrating real-time safety messages from infrastructure. These messages allow CAVs to receive immediate hazard alerts, optimize routing decisions, and make informed driving adjustments in diverse conditions.

Leveraging advancements in AI and Connected Automated Vehicles (CAVs), the EU-H2020-funded ESRIUM project aims to revolutionize European road safety and efficiency. ESRIUM develops high-precision digital maps that capture real-time road surface conditions, documenting deterioration accurately. Connected vehicles receive tailored routing and driving advice, effectively reducing road wear, decreasing maintenance requirements, and enhancing driving comfort \cite{Rudigier2022}. Building on ESRIUM’s outcomes, the ESERCOM-D project \cite{esercomd} emphasizes standardization by providing a comprehensive service designed to promote greener, smarter road usage, optimize maintenance activities, and boost road safety. Integrating multimodal large language models (MLLMs) with Vehicle-to-Anything (V2X) communications, ESERCOM-D seeks to significantly enhance traffic hazard management, increase CAV maneuverability, and ensure ongoing traffic monitoring, road condition assessments, and accident detection.

This work addresses limitations of traditional methods reliant on manual monitoring and explores the potential of infrastructure-based vision sensors and generative AI for enhanced traffic accident detection and description. By leveraging MLLMs, specifically Gemini 1.5\cite{team2024gemini}, Gemini 2.0 \cite{Gemini2}, Gemma 3 \cite{team2025gemma} and Pixtral\cite{mistralai2024pixtral}, the study evaluates zero-shot performance in identifying accidents from images. Complementary technologies such as YOLO for object detection, Deep SORT for tracking, and Segment Anything for segmentation are integrated to assist in detection and participant recognition.

\textit{\textbf{Contributions of this paper}:} Our main objectives in this paper are: (1) Evaluating zero-shot capabilities of MLLMs for accident detection using the DeepAccident dataset generated in CARLA; (2) Investigating the effects of enhanced prompts generated using computer vision techniques (YOLO, SAM, Deep SORT) on accident detection performance; (3) Comparing performance differences between four different MLLMs. In this respect, our quantitative analysis demonstrates the robust potential of MLLMs to accurately detect and describe diverse traffic accidents across various environmental conditions. Furthermore, we summarize multiple metrics to assess the explainability of MLLM-generated accident descriptions, identifying cosine similarity computed via Sentence Transformers and Word2Vec as the most effective measure of semantic similarity between model outputs and ground truth.
An unexpected finding was that enhanced prompts incorporating outputs from computer vision algorithms did not necessarily improve MLLMs' accident detection performance. We hypothesize that these enhancements might introduce irrelevant noise, potentially due to MLLMs not being trained with similar additional visual data.

The rest of this paper is structured as follows: Section~\ref{sec:rw} summarizes previous approaches and available datasets for infrastructure-based traffic accident detection, both with and without the use of Multimodal Large Language Models (MLLMs). Section~\ref{sec:PS} provides the problem statement, and Section~\ref{sec:our_approach} describes our approach to addressing it. Our findings are presented in Section~\ref{sec:experiments}. Finally, we summarize and reflect on our work in Section~\ref{sec:CO}.

\section{Related Work} \label{sec:rw}


Vision-based accident detection methods can be categorized by their model architectures into several categories, including both frame-level and object-centric approaches \cite{fang2023vision}. Frame-level accident detection relies on extracting features from individual frames and requires annotating accident windows in traffic recordings. You and Han \cite{you2020traffic} propose a dataset for traffic understanding, where accident windows are detected by classifying feature embeddings derived from video frames using a 3D-CNN model.
In contrast, methods that emphasize object detection and tracking aim to identify accident participants by detecting and localizing objects in images and monitoring their movement over time. Various approaches have used the YOLO detector \cite{7780460} previously, among which are YOLOv4 \cite{ghahremannezhad2022real} and YOLOv5 \cite{xia2022research}. Karim et al. \cite{karim2024visual} employed YOLOv8 to  detect vehicles and accidents in real time from surveillance video. They integrated Deep-SORT \cite{wojke2017simple} to track vehicle movement across frames, supporting temporal analysis for accident detection. While 3D-CNN based classification and YOLO with Deep-SORT tracking effectively pinpoint accident participants, they sometimes fall short in capturing the broader context needed for a more thorough analysis of accident scenarios.

The growth of artificial intelligence has increased the demand for extensive datasets to train models for specific applications, including accident detection. Consequently, numerous datasets have been developed to support research in this area. Yao et al. \cite{yao2022dota} propose an unsupervised traffic anomaly detection framework that employs future object localization and a novel  spatial-temporal area under curve (STAUC) on the richly annotated DoTA dataset to effectively identify abnormal events in dynamic, egocentric driving videos.
However, since the DoTA dataset primarily comprises dashcam video recordings with highly dynamic perspectives, it may not be ideally suited for applications requiring fixed, infrastructure-based scenes for accident detection.

Lee et al. \cite{lee2017crash} introduced GTACrash, a synthetic dataset derived from GTA V that enables training CNN-based collision prediction algorithms, and demonstrated that synthetic data is a valuable alternative to scarce real-world accident data, although the approach suffered from limited accident diversity and realism.
DeepAccident \cite{wang2024deepaccident} is a large-scale, simulator-generated V2X dataset for autonomous driving, developed using the CARLA simulator \cite{dosovitskiy2017carla}. It offers a wide variety of collision scenarios, accompanied by comprehensive sensor and annotation data from 6 infrastructure cameras, making it a robust resource for accident prediction and accident detection research.

ConnectGPT \cite{tong2024connectgpt} introduces a pipeline that integrates GPT-4 with infrastructure cameras and V2X communication to automatically analyze traffic conditions and generate standardized C-ITS messages, thereby enhancing real-time incident detection and road safety. Authors demonstrated its potential through practical experiments on a small dataset, highlighting improved responsiveness and reduced manual intervention in managing traffic hazards. However, its reliance solely on GPT-4 without evaluating alternative multimodal LLMs may limit its broader applicability.

AccidentGPT \cite{wu2024accidentgpt} presented a framework that combines V2X environmental perception with GPT-based reasoning to enable real-time accident prediction, prevention, and post-accident analysis. By fusing multi-sensor data and using collaborative perception techniques, the system delivers proactive alerts and detailed analyses of accident causation, aiding traffic management and enforcement agencies in improving road safety. 
Although AccidentGPT provides solid scene understanding, its reliance on complex sensor fusion and modular integration may restrict its scalability and limit its practical use in infrastructure-based accident detection. Furthermore, Zhang et al. \cite{zhang2025language} introduced a framework that leverages MLLMs for structured and scalable traffic accident analysis through video classification and visual grounding, incorporating severity-based aggregation, multimodal prompts, and a tailored evaluation metric. Lohner et al. \cite{lohner2024enhancing} introduced the idea of enhancing MLLM inputs with scene graphs.

\section{Problem Statement} \label{sec:PS}
Traffic accident detection remains a challenging and critical issue for ensuring timely hazard reduction and effective emergency response. Traditional methods predominantly rely on manual monitoring or conventional computer vision techniques trained on datasets that largely represent normal, safe driving conditions.

The scarcity of annotated datasets for traffic accidents, particularly those captured by infrastructure cameras, poses a significant challenge in developing robust accident detection and descriptive systems. This limitation not only restricts the training of advanced models but also hampers the comprehensive analysis of accident scenarios. To address this gap, we utilize the DeepAccident dataset—a synthetic dataset generated using the CARLA simulator within a controlled environment. Although the dataset includes preliminary annotations, these labels are insufficient for detailed analysis. Therefore, we have augmented the dataset with additional annotations, including frame-level accident markers, contextual scene descriptions, object involvement details, and justifications for classifying specific frames as accidents. 
Prior studies have shown that conventional computer vision algorithms, such as YOLO for object detection and Deep SORT for tracking, can effectively identify accidents. However, these methods typically fall short in providing a rich environmental context. In our work, we investigate the reasoning capabilities of multimodal large language models (MLLMs), specifically the vanilla models Gemini 1.5\cite{team2024gemini}, Gemini 2.0 \cite{Gemini2}, Gemma 3 \cite{team2025gemma} and Pixtral Large \cite{mistralai2024pixtral}, under various weather conditions using the DeepAccident dataset. 

\section{Our Approach}\label{sec:our_approach}

\subsection{Architecture Visualization}
\begin{figure}[htbp]
    \centering
    \includegraphics[width=0.9\columnwidth]{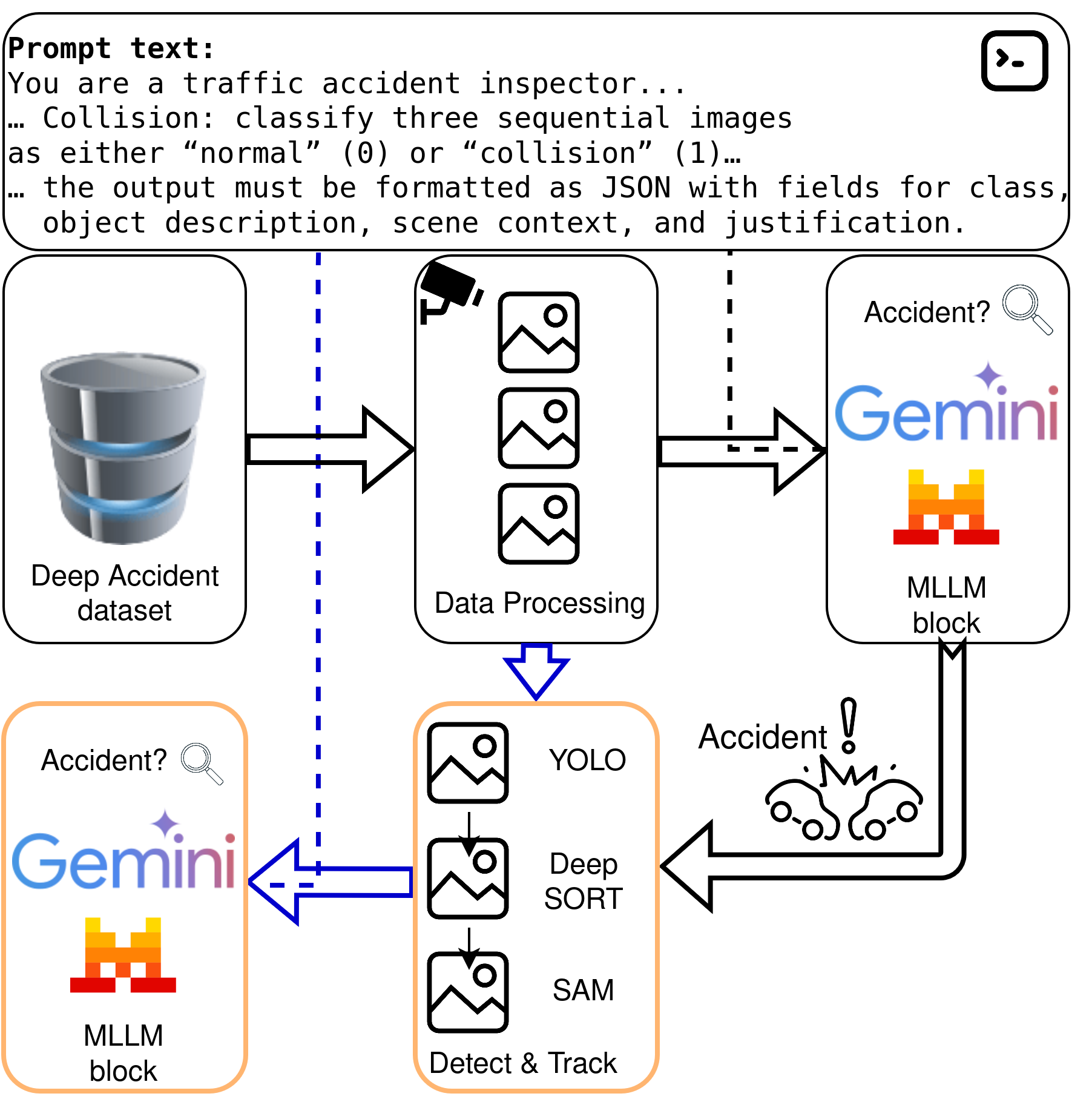}
    \caption{Accident detection framework visualization.}
    \label{fig:paper_diagram}
\end{figure}


Our approach leverages these models' ability to process visual data and generate structured textual descriptions that include:

\begin{itemize}
\item \textbf{Accident Classification}: A binary indication of accident presence in each frame.
\item \textbf{Scene Context}: Detailed descriptions of the scene, with a focus on weather and driving conditions.
\item \textbf{Justification}: Explanations supporting the classification decision.
\end{itemize}

In addition, we incorporate state-of-the-art techniques by integrating YOLO for object detection, Deep SORT for multi-object tracking, and the Segment Anything Model (SAM) for extracting object contours. This multimodal framework enriches the accident detection process by combining robust visual analysis with contextual reasoning, ultimately paving the way for advancements in automated traffic monitoring and accident response systems.

Our framework is illustrated in Figure~\ref{fig:paper_diagram}. The DeepAccident dataset is first loaded and preprocessed, as it contains additional modalities such as LiDAR data and vehicle-view recordings, which are not required for our analysis. The primary objective of the Data Processing block (Section~\ref{subsec: DP}) is to extract infrastructure camera frames for a given scenario. In addition to these three consecutive frames, a textual task description is also provided to the model. If an accident is detected, we apply an \textcolor{orange}{object detection and tracking algorithm} to generate \textcolor{orange}{enhanced prompts}, as described in Sections~\ref{subsec: OD&TM} and~\ref{subsec: EP}. The \textcolor{orange}{orange} blocks represent visual prompts.

\subsection{Data Processing}\label{subsec: DP}
\begin{figure*}[htbp]
    \centering
    \includegraphics[width=\columnwidth * 2]{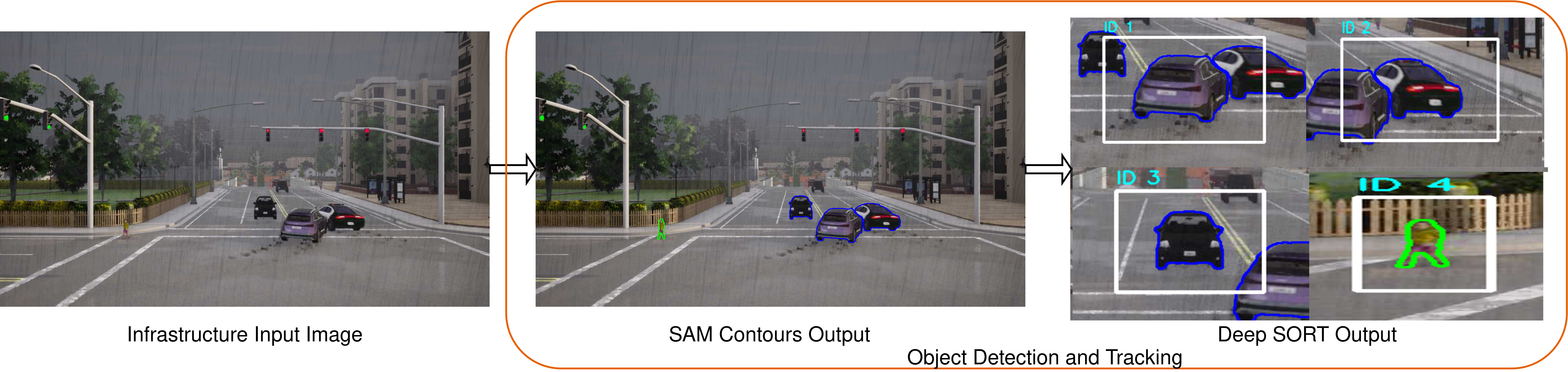}
    \caption{We apply a \textcolor{orange}{object detection and tracking algorithms} to generate enhanced prompts}
    \label{fig:ODandTM_img}
\end{figure*}
DeepAccident is an open-source dataset that contains a large-scale collection of synthetic accident scenarios generated using the CARLA simulator and features a pioneering V2X dataset specifically developed for V2X applications. The dataset encompasses a diverse range of accident scenarios across various road types, weather conditions, and times of day, focusing on 12 distinct intersection accident types—such as running a red light, unexpected left turns, and turn conflicts. Each scenario includes five agents: two vehicles with overlapping trajectories, two following vehicles, and one infrastructure unit. Every agent is outfitted with six RGB cameras and a LiDAR sensor, with labels provided in the LiDAR coordinate frame. Since our analysis concentrates on the infrastructure’s perspective, only those images are loaded and examined. Harada et al. \cite{harada2025traffic} identified challenges in the dataset, including the absence of a unified temporal annotation standard and the mixture of ego-vehicle-involved with non-involved accidents. To address these issues, we manually reviewed each scenario and enriched the dataset with additional annotations—providing scene context descriptions, detailed object information, and a justification that offers a concise scenario summary and, when applicable, details the accident and its cause. The analysis was conducted on 100 scenarios from the dataset. Although the dataset contains more scenarios, we selected 100 scenarios distributed over 12 different types (e.g., running red lights, left turns against traffic, etc.), with an equal split of scenarios with and without accidents. This selection was made because some scenarios had unclear labeling and required additional manual descriptions.


\subsection{Object Detection and Tracking Module}\label{subsec: OD&TM}


This accident detection method combines frame-level and object-centric approaches. First, objects in the image are detected and recognized. Once their bounding boxes have been successfully identified, the objects are tracked across frames and assigned unique IDs. In the final stage, the visibility of these objects is enhanced by highlighting their contours.

\subsubsection{YOLO}

In order to efficiently detect and classify vehicles and pedestrians on the street, we are deploying the YOLOv8l model \cite{ultralytics2025}. YOLO processes images in a single pass, allowing for fast and real-time detection. It has an ability to predict both bounding boxes and class probabilities simultaneously, making it perfect for this task. Bounding boxes detected by YOLO are subsequently used as inputs for SAM and DeepSORT. Although YOLO can identify numerous classes, this work focuses on the following: person, bicycle, car, motorcycle, bus, train, and truck.

\subsubsection{Deep SORT}
Simple Online and Realtime Tracking with a Deep Association Metric (Deep SORT) is used to link detected objects across video frames. It improves upon the traditional SORT \cite{bewley2016simple} algorithm by integrating deep appearance features from a CNN and employing the Hungarian algorithm to optimally assign detections to predicted tracks. This combination not only enhances tracking robustness but also allows the system to maintain tracks over multiple frames, effectively handling occlusions and missed detections. The algorithm takes the bounding boxes previously detected by the YOLO model as input, meaning that its performance is directly influenced by YOLO's efficiency. Additionally, Deep SORT uses Kalman filtering to predict an object’s next state based on its past positions, with tracking currently performed over three frames, although the implementation can support more frames if needed.

\subsubsection{SAM}
The Segment Anything Model (SAM) is a groundbreaking computer vision framework developed by Meta, capable of performing high-quality, zero-shot image segmentation on virtually any object. This module operates on input images previously processed by YOLO and DeepSORT. Segmentation is performed only on objects that have been successfully detected and classified by YOLO and tracked by DeepSORT across at least three frames. SAM is then used to extract the contours of these tracked objects within the scene.

\subsection{MLLM Block}
We leverage the robust capabilities of multimodal large language models (MLLMs) for zero-shot accident detection. All four models excel at processing multimodal inputs—including images from infrastructure cameras. Importantly, these vanilla models have not been pre-trained on any task-specific datasets. Our study evaluates their effectiveness in detecting accidents and contextualizing the surrounding environment.
\begin{table}[!t]
\renewcommand{\arraystretch}{1.3} 
\centering
\caption{Similarity Metrics Comparison in the example of the challenge.}
\label{tab:similarity}
\begin{tabular}{lcc}
\hline
\bfseries Metric & \bfseries S1 vs. S2 & \bfseries S1 vs. S3 \\
\hline
BLEU                                & 0.46 & 0.16 \\
ROUGE                               & 0.75 & 0.58 \\
Word2Vec Cosine Similarity          & 0.86 & 0.85 \\
Sentence Transformers Cosine Similarity & 0.93 & 0.84 \\
\hline
\end{tabular}
\end{table}

Each request to the model is composed of up to three consecutive frames accompanied by a textual component. The frames are extracted from the DeepAccident dataset, where each scenario comprises between 45 and 120 frames. At this stage, the images are used without any preprocessing. The prompt is specifically designed to instruct the model to classify a traffic event based on three sequential images. The classification task is to determine whether the event is “normal” (indicating no abnormal behaviour) or a “collision” (an accident involving an impact between road users).

In this context, the model acts as a traffic accident inspector, analysing the images to decide if the event demonstrates normal behaviour or a collision. Any event in which vehicles, pedestrians, or objects come into direct contact is classified as a collision. The prompt mandates that the response be provided in a structured JSON format, with collisions represented as 1 (collision) and non-collisions as 0 (normal). Furthermore, the model must supply a description of the environment, weather conditions, and overall scene context. It is also required to provide a justification for the classification decision and to identify and briefly describe the objects present in the scene. This detailed description will later be evaluated against the ground truth data. The two models used in accident detection will be introduced as follows: 

\subsubsection{Gemini Models}  Gemini 1.5 is an MLLM developed by Google DeepMind that uses a sparse mixture of experts (MoE)\cite{miller1996mixture} architecture and knowledge distillation to deliver rapid and cost-effective performance. It supports extended context windows of up to one million tokens, making it ideal for real-time applications such as interactive chatbots and long-document analysis while processing text, images, audio, and video seamlessly. Gemini 2.0, Google DeepMind’s flagship multimodal model, introduces native image and audio generation, faster tool invocation, and enhanced multi-step reasoning.

\subsubsection{Pixtral Large} This is a 124B open-weights multimodal model from Mistral AI with the ability to understand documents, diagrams, and images.  It combines a robust text decoder with an advanced vision encoder and can handle up to 128,000 tokens per inference cycle to excel in tasks like document analysis and visual question answering, which makes it perfectly suitable for our research analysis.

\subsubsection{Gemma 3} This is the latest lightweight, open-source model from Google, and the first in its family to feature vision capabilities, supporting long input sequences of up to 128K tokens on consumer-level hardware. To handle this long context efficiently, it uses an interleaved global-local attention mechanism and is trained via knowledge distillation from larger models to enhance its quality.

\subsection{Enhanced Prompts}\label{subsec: EP}
Enhanced prompts maintain the original prompt text described above while incorporating an adjusted visual component. First, we load the frames classified as an accident (classification 1) and process them in batches of three using YOLO for object detection. Next, DeepSORT assigns unique identifiers to the detections from YOLO. These processed prompts are then forwarded to the SAM model for contour extraction. Pedestrians are colored in \textcolor{green}{green} and other traffic participants in \textcolor{blue}{blue}. This is also shown in Figure~\ref{fig:ODandTM_img}. Finally, we have the model evaluate these prompts to provide again a classification, justification, scene context, and a detailed description of the objects involved.

\section{Experiments and Discussion} \label{sec:experiments}

\begin{table}[h]
    \caption{Overall Average Metrics Across All Scenarios}
    \centering
    \renewcommand{\arraystretch}{1.3} 
    \begin{tabular}{lcc}
        \hline
        \textbf{Metric} & \textbf{Gemini} & \textbf{Pixtral} \\
        \hline
         \textbf{Cosine (word2vec) }       & 0.56 & 0.58 \\
         \textbf{Cosine (sentence transformers)} & 0.48 & 0.52 \\
        \hline
    \end{tabular}
    \label{tab:overall_metrics}
\end{table}

\subsection{Metrics}
In this section, we survey the metrics employed for evaluating traffic accident detection and discuss their effectiveness in this specific case. Fang and colleagues \cite{fang2023vision} highlight Precision, Recall, and F1-score as the primary metrics for visual traffic accident detection.  Additionally, we assess model performance using Accuracy, BLEU, ROUGE, along with cosine similarity measures derived from Word2Vec and Sentence Transformer embeddings. 

\textbf{Precision} measures the proportion of instances predicted as traffic accidents that are indeed correct. \textbf{Recall} quantifies the proportion of all actual traffic accidents that the model successfully identifies. \textbf{F1 score} is the harmonic mean of precision and recall, which provides a balanced measure of the model’s performance.

\textbf{BLEU} and \textbf{ROUGE} are standard choices for evaluating text quality in natural language processing. However, as highlighted in \cite{zhang2025language}, these metrics primarily assess surface-level similarities between the predicted and reference sentences. This emphasis on superficial matching is particularly problematic in domains like traffic accident analysis, where even a single misinterpreted word can dramatically alter the interpretation of an event. 

An example of this challenge can be seen through the following case: We define three sentences that convey the same underlying meaning but are expressed using different words and orders: 

\begin{description}
    \item[(S1)] The accident occurred when a vehicle lost control, likely due to driver inattention, and collided with a roadside barrier.
    \item[(S2)] A vehicle lost control, probably due to driver inattention, and struck a roadside barrier, resulting in the accident.
    \item[(S3)] Because the driver was inattentive, the vehicle lost control, veered off course, and struck a roadside barrier, which led to the accident.
\end{description}

Here, the ground truth (S1) is compared with two variations (S2 and S3). To address the challenge mentioned above, we incorporate semantic similarity measures. \textbf{Word2Vec} \cite{mikolov2013distributed} transforms individual words into continuous vector representations based on their contextual usage, enabling us to gauge similarity via the proximity of these vectors. \textbf{Sentence Transformers} \cite{reimers2019sentence, wang2020minilm} extend this approach by encoding entire sentences into context-aware embeddings, thereby capturing richer semantic nuances. Cosine similarity, defined in Equation~\ref{eq:cosine_similarity}, is then used to quantify the similarity between the vector representations, with scores near 1 indicating high similarity. 

\begin{equation}
\text{cosine\_similarity}(\mathbf{A}, \mathbf{B}) = \frac{\mathbf{A} \cdot \mathbf{B}}{\|\mathbf{A}\| \|\mathbf{B}\|}
\label{eq:cosine_similarity}
\end{equation}

Table~\ref{tab:similarity} summarizes the scores across four metrics in the example of the challenge. Cosine similarity from both Word2Vec and Sentence Transformers shows more stable results compared to the variability seen with BLEU and ROUGE. Consequently, the cosine similarity metrics from Word2Vec and Sentence Transformers are employed in the subsequent performance evaluations of MLLMs and architectural frameworks due to their consistent and robust performance.

\subsection{Performance Evaluation}

This section assesses the performance of four models: Gemini 1.5, Gemini 2.0, Pixtral, and Gemma 3. We evaluate these models with and without enhanced prompts, based on the metrics established previously.

\begin{figure}[htbp]
    \centering
    \subfigure[Object Description]{
        \includegraphics[width=1.0\columnwidth]{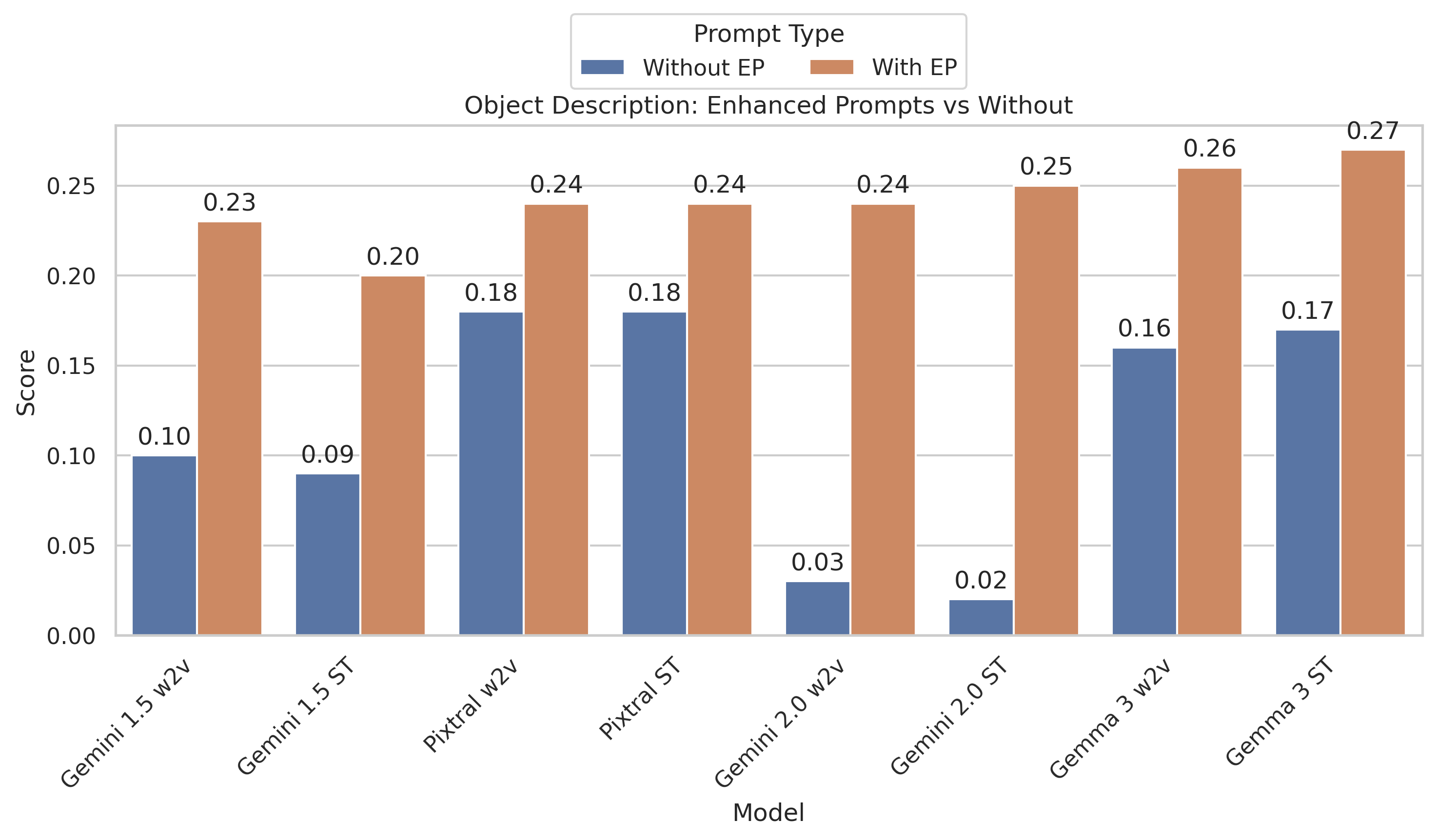}
        \label{fig:object_description}
    }
    \subfigure[Scene Context]{
        \includegraphics[width=1.0\columnwidth]{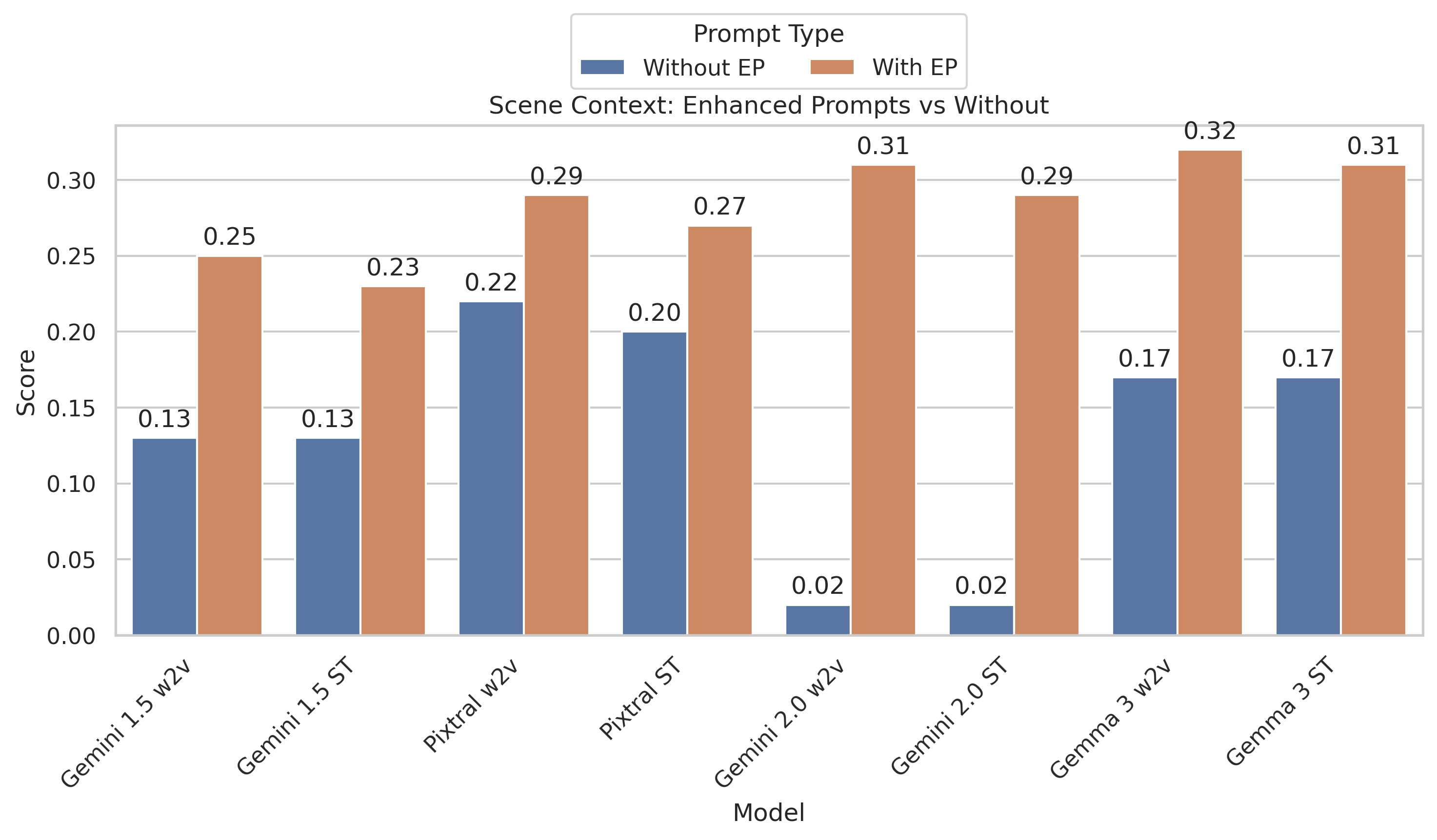}
        \label{fig:scene_context}
    }
    \subfigure[Justification]{
        \includegraphics[width=1.0\columnwidth]{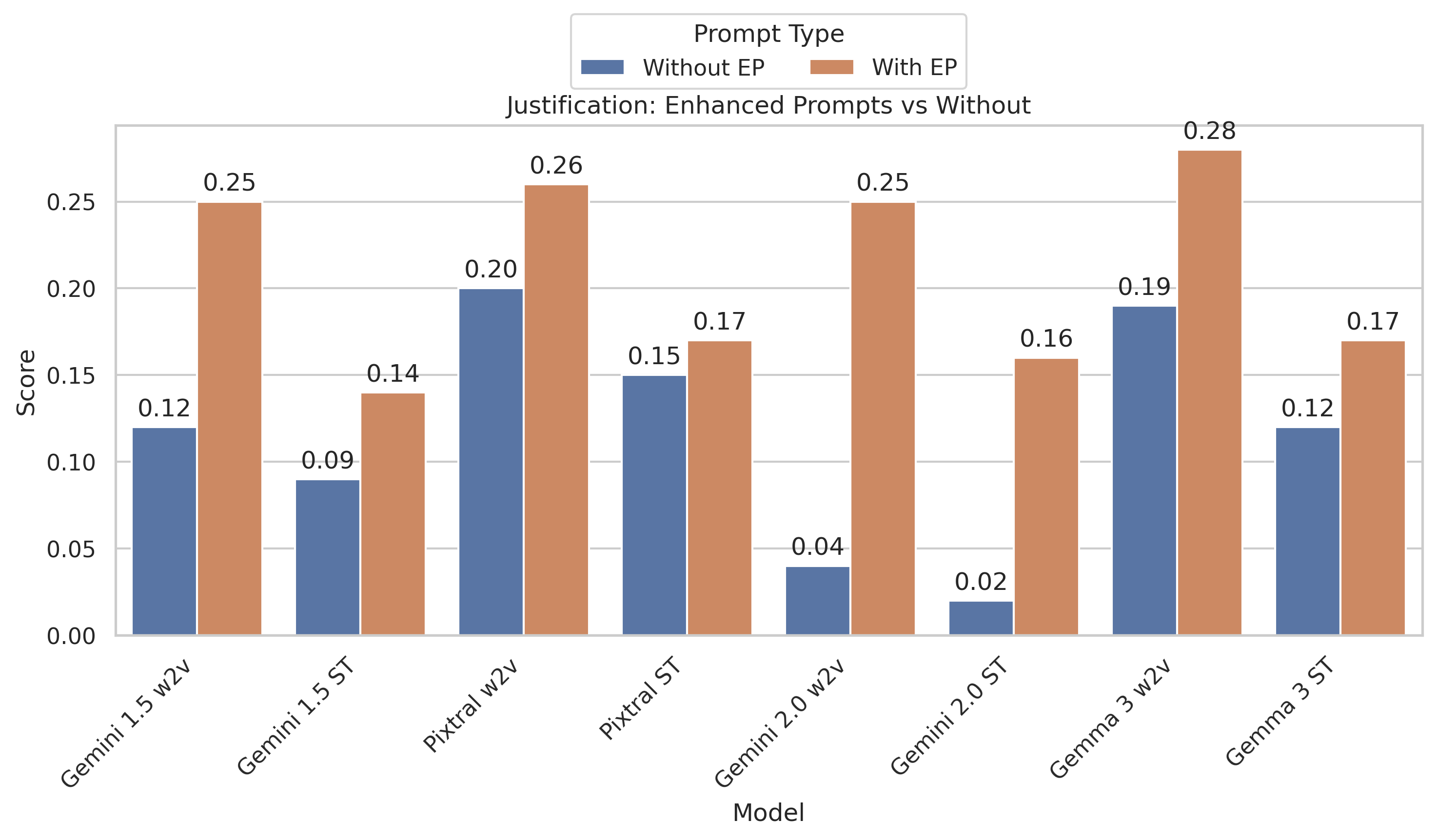}
        \label{fig:justification}
    }
    \caption{Performance metrics for models with and without enhanced prompts (EP) across three tasks.}
    \label{fig:combined_performance}
\end{figure}

Figure~\ref{fig:combined_performance} confirms that all models achieve higher scores without enhanced prompts. Although enhanced prompts were intended to improve output quality, the results show that all models generated more accurate answers without them across all three categories—Justification, Scene Context, and Object Description.
Table~\ref{tab:overall_metrics} shows the average performance of the Gemini and Pixtral models across all scenarios, using cosine similarity (via word2vec and sentence transformers) to assess semantic alignment. Overall, Pixtral slightly outperforms Gemini in cosine similarity scores, suggesting that its generated descriptions are more closely aligned with the ground truth.

\begin{figure}[htbp]
    \centering
    \includegraphics[width=1.0\columnwidth]{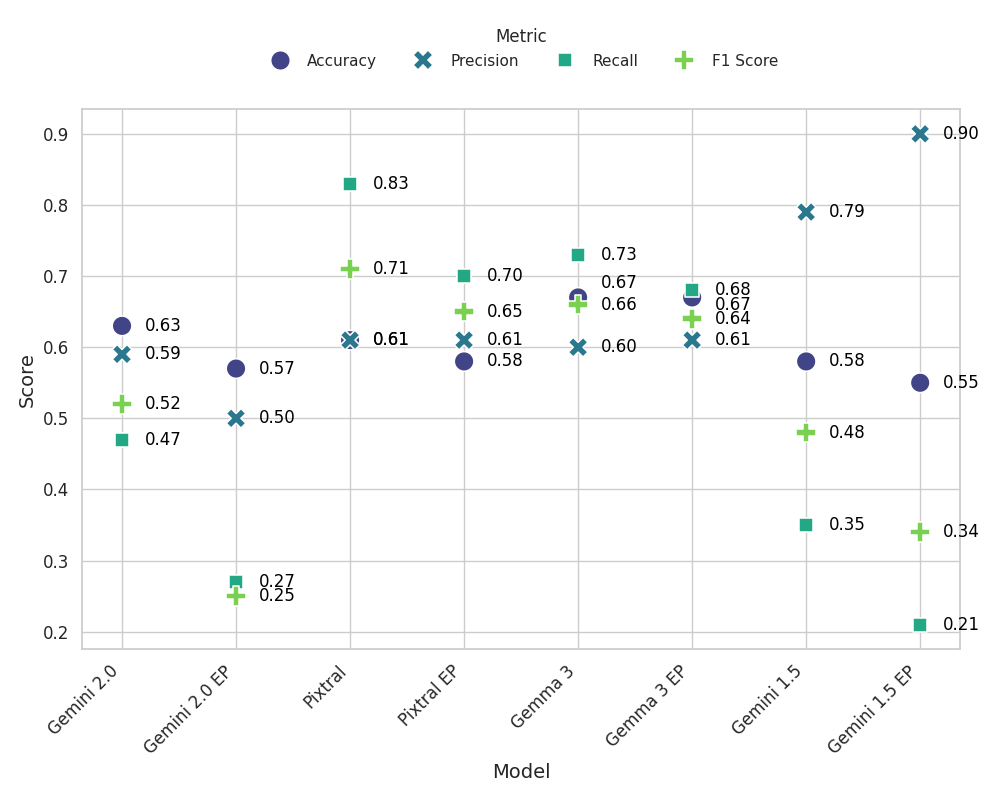}
    \caption{Performance metrics for models with and without enhanced prompts (EP).}
    \label{fig:accident_detectionpaper_diagram}
\end{figure}

Figure~\ref{fig:accident_detectionpaper_diagram} summarizes performance across models with and without enhanced prompts. Enhanced prompts generally increased precision but reduced recall and F1, key metrics for our use case. Gemini 1.5, for example, saw precision rise to 90\% with EPs, but recall dropped sharply, leading to a 14 percentage-point F1 loss. Gemini 2.0 showed an even steeper decline. Pixtral remained the top performer with 83\% recall and 71\% F1, while Gemma 3 showed the most balanced trade-off with minimal metric changes.

\section{Conclusion and Outlook} \label{sec:CO}
Our work investigates the ability of MLLMs to detect and accurately describe diverse traffic accidents across a range of scenarios using the DeepAccident dataset, employing various metrics to evaluate their performance. We employ and compare the Gemini 1.5 and 2.0, Gemma 3 and Pixtral MLLMs to perform this task. Additionally, we evaluate their performance on the same set of scenarios augmented with detection data from advanced detection algorithms. Finally, we briefly discuss and present different semantic measurements for assessing their descriptions of traffic situations. We show that MLLMs are useful in this context because they effectively integrate visual and textual information, enabling a more comprehensive understanding of complex traffic scenarios. We find complementary strengths across the four models: Gemini 1.5, Gemini 2.0, Pixtral, and Gemma 3. Pixtral delivers the best overall performance with the highest recall and F1 scores, showing strong detection capabilities. Gemini 1.5 and Gemini 2.0 improve in precision when using enhanced prompts but experience notable drops in recall and F1. Gemma 3 maintains the most stable performance, balancing precision and recall with minimal variation. These results highlight the trade-offs between completeness and accuracy among the models.

Our experiments also indicate that integrating enhanced prompts with YOLO, SAM, and DeepSORT introduces additional detection information that, when not accompanied by specialized fine-tuning, significantly degrades performance across all evaluated metrics. This decrease likely arises because the raw detection outputs introduce noise that the base models are not optimized to handle, thereby impeding their ability to accurately interpret accident scenarios. 
A recent study found a similar effect. Jiao et al. \cite{jiao2024enhancing} observed that simply adding detection data using a "Training-free Infusion" method actually lowered performance on several tests. This suggests that MLLMs have a hard time using this extra information correctly without additional training. In other words, more training is needed so the models can ignore unnecessary details and make better use of detection-enhanced prompts for spotting traffic accidents.
Our future work will focus on improving the performance of the proposed architecture and further analysis of the results to better understand the effect of enhanced prompts. To bridge the sim-to-real gap, we plan to deploy our solution on a real-world traffic surveillance system. The extended work will include the detection of not only traffic accidents but also a broader range of traffic anomalies. Ultimately, our efforts will involve fine-tuning the models to enhance their ability to detect anomalies and assist human traffic supervisors.

\section*{Acknowledgment}
The work was supported by the project ESERCOM-D. The Project ESERCOM-D is funded by the European Union under grant agreement No 101180176. Views and opinions expressed are however those of the author(s) only and do not necessarily reflect those of the European Union or European Union Agency for the Space Programme (EUSPA). Neither the European Union nor the granting authority can be held responsible for them.
The publication was written at Virtual Vehicle Research GmbH in Graz and partially funded within the COMET K2 Competence Centers for Excellent Technologies by the Austrian Federal Ministry for Innovation, Mobility and Infrastructure (BMIMI), Austrian Federal Ministry for Economy, Energy and Tourism (BMWET), the Province of Styria (Dept. 12) and the Styrian Business Promotion Agency (SFG). The Austrian Research Promotion Agency (FFG) has been authorised for the programme management.

\bibliographystyle{IEEEtran}

\bibliography{ref.bib}

\begin{thebibliography}{10}
\providecommand{\url}[1]{#1}
\csname url@samestyle\endcsname
\providecommand{\newblock}{\relax}
\providecommand{\bibinfo}[2]{#2}
\providecommand{\BIBentrySTDinterwordspacing}{\spaceskip=0pt\relax}
\providecommand{\BIBentryALTinterwordstretchfactor}{4}
\providecommand{\BIBentryALTinterwordspacing}{\spaceskip=\fontdimen2\font plus
\BIBentryALTinterwordstretchfactor\fontdimen3\font minus
  \fontdimen4\font\relax}
\providecommand{\BIBforeignlanguage}[2]{{%
\expandafter\ifx\csname l@#1\endcsname\relax
\typeout{** WARNING: IEEEtran.bst: No hyphenation pattern has been}%
\typeout{** loaded for the language `#1'. Using the pattern for}%
\typeout{** the default language instead.}%
\else
\language=\csname l@#1\endcsname
\fi
#2}}
\providecommand{\BIBdecl}{\relax}
\BIBdecl

\bibitem{Rudigier2022}
M.~Rudigier, S.~Solmaz, G.~Nestlinger, and K.~Tong, ``Development, verification
  and kpi analysis of infrastructure-assisted trajectory planners,'' in
  \emph{2022 International Conference on Connected Vehicle and Expo (ICCVE)},
  2022, pp. 1--6.

\bibitem{esercomd}
``{ESERCOM-D},'' \url{https://esercomd.eu/}, accessed: 27 March 2025.

\bibitem{team2024gemini}
G.~Team, P.~Georgiev, V.~I. Lei, R.~Burnell, L.~Bai, A.~Gulati, G.~Tanzer,
  D.~Vincent, Z.~Pan, S.~Wang \emph{et~al.}, ``Gemini 1.5: Unlocking multimodal
  understanding across millions of tokens of context,'' \emph{arXiv preprint
  arXiv:2403.05530}, 2024.

\bibitem{Gemini2}
\BIBentryALTinterwordspacing
G.~2.0, ``Gemini2,'' Google Deep Mind, 2024, accessed: April 27, 2025.
  [Online]. Available:
  \url{https://blog.google/technology/google-deepmind/google-gemini-ai-update-december-2024/\#ceo-message}
\BIBentrySTDinterwordspacing

\bibitem{team2025gemma}
G.~Team, A.~Kamath, J.~Ferret, S.~Pathak, N.~Vieillard, R.~Merhej, S.~Perrin,
  T.~Matejovicova, A.~Ram{\'e}, M.~Rivi{\`e}re \emph{et~al.}, ``Gemma 3
  technical report,'' \emph{arXiv preprint arXiv:2503.19786}, 2025.

\bibitem{mistralai2024pixtral}
\BIBentryALTinterwordspacing
M.~AI, ``Pixtral-large-instruct-2411,'' Hugging Face, 2024, accessed: April 09,
  2025. [Online]. Available:
  \url{https://huggingface.co/mistralai/Pixtral-Large-Instruct-2411}
\BIBentrySTDinterwordspacing

\bibitem{fang2023vision}
J.~Fang, J.~Qiao, J.~Xue, and Z.~Li, ``Vision-based traffic accident detection
  and anticipation: A survey,'' \emph{IEEE Transactions on Circuits and Systems
  for Video Technology}, vol.~34, no.~4, pp. 1983--1999, 2023.

\bibitem{you2020traffic}
T.~You and B.~Han, ``Traffic accident benchmark for causality recognition,'' in
  \emph{Computer Vision--ECCV 2020: 16th European Conference, Glasgow, UK,
  August 23--28, 2020, Proceedings, Part VII 16}.\hskip 1em plus 0.5em minus
  0.4em\relax Springer, 2020, pp. 540--556.

\bibitem{7780460}
J.~Redmon, S.~Divvala, R.~Girshick, and A.~Farhadi, ``You only look once:
  Unified, real-time object detection,'' in \emph{2016 IEEE Conference on
  Computer Vision and Pattern Recognition (CVPR)}, 2016, pp. 779--788.

\bibitem{ghahremannezhad2022real}
H.~Ghahremannezhad, H.~Shi, and C.~Liu, ``Real-time accident detection in
  traffic surveillance using deep learning,'' in \emph{2022 IEEE international
  conference on imaging systems and techniques (IST)}.\hskip 1em plus 0.5em
  minus 0.4em\relax IEEE, 2022, pp. 1--6.

\bibitem{xia2022research}
Z.~Xia, J.~Gong, H.~Yu, W.~Ren, and J.~Wang, ``Research on urban traffic
  incident detection based on vehicle cameras,'' \emph{Future Internet},
  vol.~14, no.~8, p. 227, 2022.

\bibitem{karim2024visual}
A.~Karim, M.~A. Raza, Y.~Z. Alharthi, G.~Abbas, S.~Othmen, M.~S. Hossain,
  A.~Nahar, and P.~Mercorelli, ``Visual detection of traffic incident through
  automatic monitoring of vehicle activities,'' \emph{World Electric Vehicle
  Journal}, vol.~15, no.~9, p. 382, 2024.

\bibitem{wojke2017simple}
N.~Wojke, A.~Bewley, and D.~Paulus, ``Simple online and realtime tracking with
  a deep association metric,'' in \emph{2017 IEEE international conference on
  image processing (ICIP)}.\hskip 1em plus 0.5em minus 0.4em\relax IEEE, 2017,
  pp. 3645--3649.

\bibitem{yao2022dota}
Y.~Yao, X.~Wang, M.~Xu, Z.~Pu, Y.~Wang, E.~Atkins, and D.~J. Crandall, ``Dota:
  Unsupervised detection of traffic anomaly in driving videos,'' \emph{IEEE
  transactions on pattern analysis and machine intelligence}, vol.~45, no.~1,
  pp. 444--459, 2022.

\bibitem{lee2017crash}
K.~Lee, H.~Kim, and C.~Suh, ``Crash to not crash: Playing video games to
  predict vehicle collisions,'' \emph{Place holder journal}, 2017.

\bibitem{wang2024deepaccident}
T.~Wang, S.~Kim, J.~Wenxuan, E.~Xie, C.~Ge, J.~Chen, Z.~Li, and P.~Luo,
  ``Deepaccident: A motion and accident prediction benchmark for v2x autonomous
  driving,'' in \emph{Proceedings of the AAAI Conference on Artificial
  Intelligence}, vol.~38, no.~6, 2024, pp. 5599--5606.

\bibitem{dosovitskiy2017carla}
A.~Dosovitskiy, G.~Ros, F.~Codevilla, A.~Lopez, and V.~Koltun, ``Carla: An open
  urban driving simulator,'' in \emph{Conference on robot learning}.\hskip 1em
  plus 0.5em minus 0.4em\relax PMLR, 2017, pp. 1--16.

\bibitem{tong2024connectgpt}
K.~Tong and S.~Solmaz, ``Connectgpt: Connect large language models with
  connected and automated vehicles,'' in \emph{2024 IEEE Intelligent Vehicles
  Symposium (IV)}.\hskip 1em plus 0.5em minus 0.4em\relax IEEE, 2024, pp.
  581--588.

\bibitem{wu2024accidentgpt}
K.~Wu, W.~Li, and X.~Xiao, ``Accidentgpt: Large multi-modal foundation model
  for traffic accident analysis,'' \emph{arXiv preprint arXiv:2401.03040},
  2024.

\bibitem{zhang2025language}
R.~Zhang, B.~Wang, J.~Zhang, Z.~Bian, C.~Feng, and K.~Ozbay, ``When language
  and vision meet road safety: leveraging multimodal large language models for
  video-based traffic accident analysis,'' \emph{arXiv preprint
  arXiv:2501.10604}, 2025.

\bibitem{lohner2024enhancing}
A.~Lohner, F.~Compagno, J.~Francis, and A.~Oltramari, ``Enhancing
  vision-language models with scene graphs for traffic accident
  understanding,'' in \emph{2024 IEEE International Automated Vehicle
  Validation Conference (IAVVC)}.\hskip 1em plus 0.5em minus 0.4em\relax IEEE,
  2024, pp. 1--7.

\bibitem{harada2025traffic}
K.~HARADA, Y.~MARUYAMA, T.~TASHIRO, and G.~OHASHI, ``Traffic accident
  prediction without object detection for single-vehicle accidents,''
  \emph{IEICE Transactions on Fundamentals of Electronics, Communications and
  Computer Sciences}, p. 2024IMP0005, 2025.

\bibitem{ultralytics2025}
\BIBentryALTinterwordspacing
{Ultralytics}. (2025) Yolov8 documentation. Accessed: 2025-03-25. [Online].
  Available: \url{https://docs.ultralytics.com/models/yolov8/}
\BIBentrySTDinterwordspacing

\bibitem{bewley2016simple}
A.~Bewley, Z.~Ge, L.~Ott, F.~Ramos, and B.~Upcroft, ``Simple online and
  realtime tracking,'' in \emph{2016 IEEE international conference on image
  processing (ICIP)}.\hskip 1em plus 0.5em minus 0.4em\relax Ieee, 2016, pp.
  3464--3468.

\bibitem{miller1996mixture}
D.~J. Miller and H.~Uyar, ``A mixture of experts classifier with learning based
  on both labelled and unlabelled data,'' \emph{Advances in neural information
  processing systems}, vol.~9, 1996.

\bibitem{mikolov2013distributed}
T.~Mikolov, I.~Sutskever, K.~Chen, G.~S. Corrado, and J.~Dean, ``Distributed
  representations of words and phrases and their compositionality,''
  \emph{Advances in neural information processing systems}, vol.~26, 2013.

\bibitem{reimers2019sentence}
N.~Reimers and I.~Gurevych, ``Sentence-bert: Sentence embeddings using siamese
  bert-networks,'' \emph{arXiv preprint arXiv:1908.10084}, 2019.

\bibitem{wang2020minilm}
W.~Wang, F.~Wei, L.~Dong, H.~Bao, N.~Yang, and M.~Zhou, ``Minilm: Deep
  self-attention distillation for task-agnostic compression of pre-trained
  transformers,'' \emph{Advances in neural information processing systems},
  vol.~33, pp. 5776--5788, 2020.

\bibitem{jiao2024enhancing}
Q.~Jiao, D.~Chen, Y.~Huang, Y.~Li, and Y.~Shen, ``Enhancing multimodal large
  language models with vision detection models: An empirical study,''
  \emph{arXiv preprint arXiv:2401.17981}, 2024.

\end{thebibliography}

\end{document}